\documentclass[11pt,a4paper]{article}

\usepackage[utf8]{inputenc}      
\usepackage[T1]{fontenc}       
\usepackage{lmodern}           
\usepackage{geometry}          
\usepackage{setspace}          
\usepackage{graphicx}          
\usepackage{caption}
\usepackage{subcaption}
\usepackage{amsmath, amssymb}  
\usepackage{booktabs}          
\usepackage{tabularx}
\usepackage{array}
\usepackage{float}
\usepackage{multirow}

\usepackage{algorithm}
\usepackage{algorithmic}

\usepackage{hyperref}          
\usepackage[round]{natbib} 

\usepackage{enumitem}          

\title{Improving Diagnostic Performance on Small and Imbalanced Datasets Using Class-Based Input Image Composition}
\author{
  HLALI Azzeddine *, Majid BEN YAKHLEF and Soulaiman EL HAZZAT\\
  Engineering Sciences Laboratory Polydisciplinary,\\ Faculty of Taza, Sidi Mohamed Ben Abdellah University (USMBA), Fes, Morocco\\
  * Correspondence: azzeddine.hlali@usmba.ac.ma
}
\date{\today}

\begin{document}

\maketitle

\begin{abstract}
    Small, imbalanced datasets and poor input image quality can lead to high false predictions rates with deep learning models. This paper introduces Class-Based Image Composition, an approach that allows us to reformulate training inputs through a fusion of multiple images of the same class into combined visual composites, named Composite Input Images (CoImg). That enhances the intra-class variance and improves the valuable information density per training sample and increases the ability of the model to distinguish between subtle disease patterns.
    Our method was evaluated on the Optical Coherence Tomography Dataset for Image-Based Deep Learning Methods (OCTDL) \citep{kulyabin2024octdl}, which contains 2,064 high-resolution optical coherence tomography (OCT) scans of the human retina, representing seven distinct diseases with a significant class imbalance.
    We constructed a perfectly class-balanced version of this dataset, named Co-OCTDL, where each scan is presented as a 3x1 layout composite image. To assess the effectiveness of this new representation, we conducted a comparative analysis between the original dataset and its variant using a VGG16 model. A fair comparison was ensured by utilizing the identical model architecture and hyperparameters for all experiments.
    The proposed approach markedly improved diagnostic results.The enhanced Dataset achieved near-perfect accuracy $(99.6\%)$ with F1-score (0.995) and AUC (0.9996), compared to a baseline model trained on raw dataset. The false prediction rate was also significantly lower, this demonstrates that the method can produce high-quality predictions even for weak datasets affected by class imbalance or small sample size.
\end{abstract}


\section{Introduction}
Deep learning models for medical image analysis are powerful, but their accuracy is often limited by training data. The two primary challenges are class imbalance, where certain pathologies are under-represented, and high visual similarity between different diseases. These issues lead to diagnostic errors \citep{anwar2018medical}, a problem exemplified in retinal imaging where distinct conditions like Age-related Macular Degeneration (AMD) and Diabetic Retinopathy (DR) can share similar visual features.

Researchers have explored various techniques to address data challenges in medical imaging. Common approaches include combining models like ResNet-50 with Random Forest \citep{hassan2023enhanced}, using transfer learning with architectures such as EfficientNet-B3 \citep{bilal2025neutrosophic}, and generating synthetic data \citep{fekri2019generating}. While these methods can mitigate class imbalance and enhance model generalization, they often have significant drawbacks. Many rely on complex architectures or external data sources and may still fail to accurately differentiate between diseases that share highly similar input raw images.

Our work tackles prediction errors caused by ambiguous or low-feature of input raw images in individual training samples. We introduce the Composite Input Image (Co-Img) approach, where we algorithmically generate structured layouts (e.g., 3x1 grids) of multiple images from the same class. This strategy is designed to amplify discriminative signals by increasing feature density and intra-class variability within each training sample, while also serving as a targeted data augmentation technique to balance under-represented classes.

We make two primary contributions. First, we propose Class-Based Input Image Composition (CB-ImgComp), a novel augmentation strategy for imbalanced medical data. Second, we validate its effectiveness on retinal OCT images, showing a significant reduction in classification errors. Our results position CB-ImgComp as a complementary and powerful tool for improving model performance on small datasets.

\section{Related works}
Deep learning in biomedical imaging has faced persistent challenges due to class imbalance, limited dataset sizes, and high inter-class visual similarity. Different strategies have been developed in the literature to mitigate these issues. We summarize them below according to four main categories.

\subsection{Class rebalancing techniques}
Imbalanced datasets are often tackled with loss re-weighting. Unified Focal Loss improves Dice/CE losses by modulating easy vs. hard pixels across medical benchmarks \citep{relatedwork_O01}. Class-Balanced Loss weights classes by effective sample numbers, widely applied to medical classification \citep{relatedwork_O02}. Asymmetric Loss down-weights easy negatives, useful for multi-label settings like chest X-rays \citep{relatedwork_O03}. Recent works also adapt Batch-Balanced Focal Loss specifically for medical classification \citep{relatedwork_O04}.

\subsection{Data augmentation \& synthetic generation}
GANs have been used to enrich small datasets, e.g., synthetic liver lesions improved CNN detection \cite{Frid-Adar2018}. Domain-driven augmentations like MixUp and CutMix show robust calibration in medical imaging \citep{Rao2023}. Diffusion models now produce realistic medical scans (MRI to CT, CBCT to CT), summarized in surveys and challenges like SynthRAD2023 \citep{Pan2024}.

\subsection{Hybrid architectures \& transfer learning}
TransUNet combines CNNs and Transformers for multi-organ segmentation \citep{chen2021transunet}, while Swin-UNet adapts hierarchical Transformers to medical tasks \citep{cao2022swin}. Transfer learning effectiveness has been questioned: Transfusion showed ImageNet pretraining offers limited gains in some medical contexts \citep{raghu2019transfusion}.

\subsection{Handling inter-class similarity}
A major challenge in retinal disease classification is the high visual overlap between conditions such as age-related macular degeneration (AMD) and diabetic retinopathy (DR), which often leads to misclassification. Recent studies have sought to mitigate this by combining segmentation with classification to better localize lesion-rich areas \citep{Li2024} and by employing attention-guided frameworks that highlight discriminative regions and improve interpretability \citep{Li_2_2024}. Reviews of AI applications in chorioretinal imaging also emphasize that while these methods reduce confusion across similar pathologies, their performance remains highly dependent on annotation quality and dataset diversity \citep{Driban2024}.
\newline\newline
\textbf{Synthesis and Gap:}\newline
Taken together, these contributions have advanced biomedical image analysis by reducing imbalance, improving diversity, and refining model architectures. However, their limitations remain evident:
\begin{itemize}[leftmargin=2em,labelwidth=1em,labelsep=0.3em,align=parleft, topsep=6pt, partopsep=6pt, parsep=0pt, itemsep=4pt]
    \item Rebalancing: may distort distributions.
    \item Augmentation: can generate unrealistic samples.
    \item Hybrid architectures: add complexity.
    \item ROI segmentation: struggles when diseases have overlapping patterns.
\end{itemize}
In contrast, our proposed CB-ImgComp approach directly enhances intra-class variability and information density at the input level by combining multiple same-class images into structured composites. Unlike classical oversampling, CB-ImgComp prevents the model from repeatedly seeing the same minority sample by varying image position with lite orientation. Unlike traditional augmentation, it preserves semantic consistency by only merging within the same class. Thus, CB-ImgComp complements and extends existing strategies as a lightweight yet effective method to reduce false positives and negatives in small, imbalanced biomedical datasets.

\section{Methodology}
\subsection{Overview of Approach}
This study proposes \textbf{Class-Based Input Image Composition} (\textbf{CB-ImgComp}),a method to enrich small and imbalanced datasets for retinal OCT classification. Instead of relying on single images, we construct structured composites of same-class samples (e.g., 3×1 layouts), which increase intra-class variability, enhance the information density per training sample, and reduce the risk of false positives/negatives caused by visual similarity across different disease classes.

\subsection{Dataset Description}
We employed the publicly available OCTDL dataset, which contains 2,064 high-resolution OCT images categorized into seven classes: Age-related Macular Degeneration (AMD), Diabetic Macular Edema (DME), Epiretinal Membrane (ERM), Retinal Artery Occlusion (RAO), Retinal Vein Occlusion (RVO), Vitreomacular Interface Disease (VID), and Normal scans.
Each image was labeled by medical experts to ensure diagnostic reliability. This dataset has been widely used in ophthalmology-related deep learning research, making it a strong benchmark for evaluating augmentation strategies \citep{kulyabin2024octdl}.

\subsection{Data Preprocessing with CB-ImgComp}
\subsubsection{Enhancing the Density of Valuable Diagnostic Information}
In addition to high inter-class similarity, model performance is often degraded by information-scarce training samples that lack strong discriminative features. These weak instances hinder effective learning.
To address this, we introduce the \textbf{CB-ImgComp}, an input-level technique that combines multiple same-class images into a single, richer sample. This method increases data diversity and information density, compelling the model to learn more robust and relevant features, thereby improving classification accuracy.
\begin{figure}[h!]
    \centering
    \includegraphics[width=0.75\textwidth]{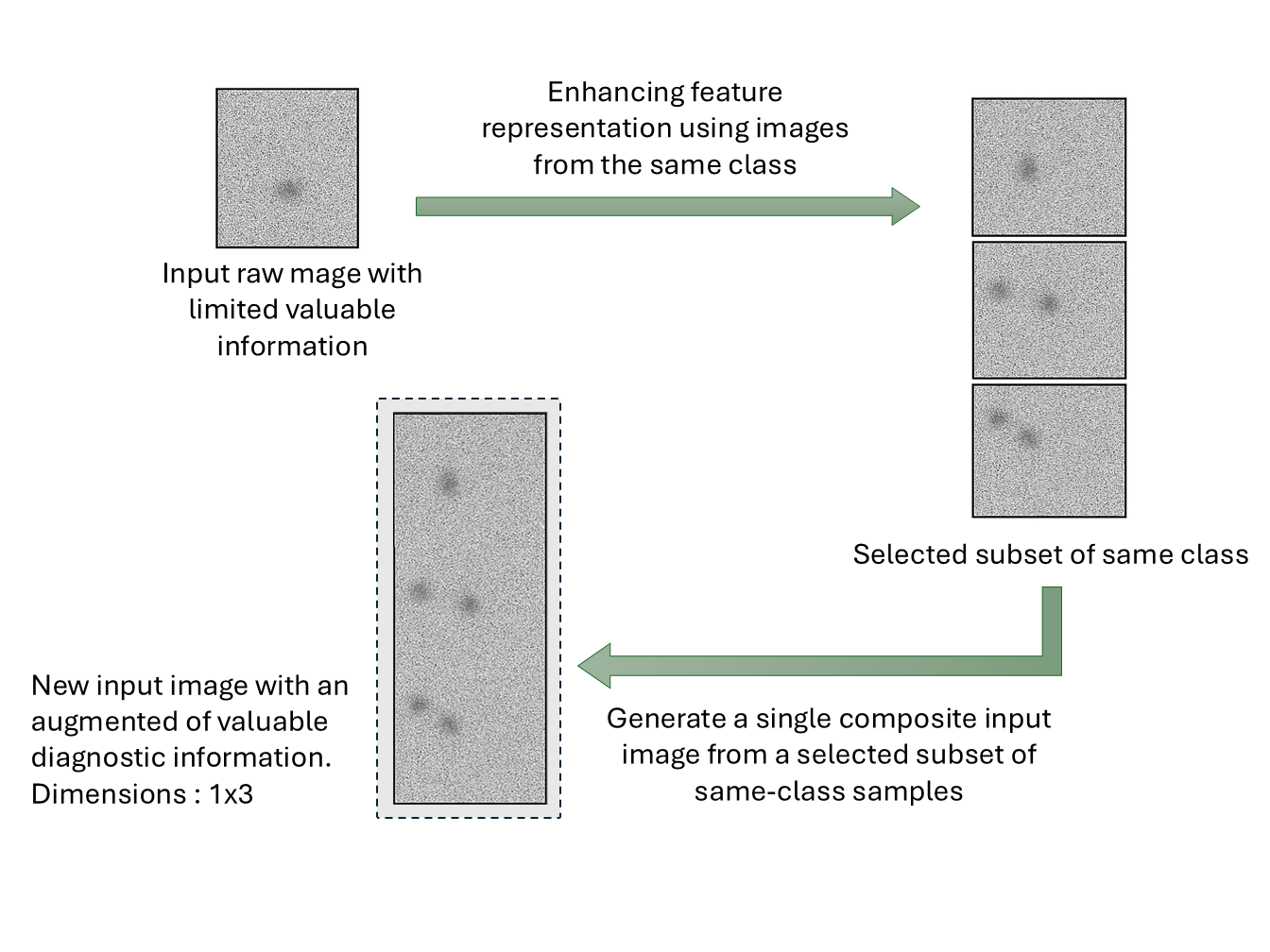}
    \caption{Constructing enhanced input images by merging same-class samples.}
    \label{fig:Constructing_enhanced_schema01}
\end{figure}

This strategy maximizing the pertinent information content of each training sample. The process creates a new representation of the original dataset called Co-OCTDL, which consists of Composite Input Images (Co-Img). The dimensionality of these Co-Img is flexible, allowing for various matrix configurations (e.g., 3×1, 2×5, 4×2) to be tailored to specific model and initial dataset requirements.
\subsubsection{Composite Input Image Generation Process}
The process of Co-Img generation passes by three steps:
\begin{enumerate}[label=\arabic*., topsep=6pt,itemsep=4pt]
    \item \textbf{Dimension Setting}: The first step is to define the format of Co-Img, which could vary from tiny configurations like 1x3, 6x12 to bigger dimensions.
    \item \textbf{Input Images Grouping}: Images are then grouped into a Co-Img by one of the following techniques:
          \begin{itemize}[leftmargin=1.4em,labelwidth=1em,labelsep=0.3em,align=parleft, topsep=6pt, partopsep=6pt, parsep=0pt, itemsep=4pt]
              \item \textbf{Class-Based Selection Function}: This approach picks images from a given class by grouping them into combinations without repetition.
              \item \textbf{Class-Based Similarity Function Selection}: This approach uses a selection function depending on the degree of similarity—either high or low—within a class domain. Every Co-Img includes images with a high/low difference of similarity.
              \item \textbf{Class-Based Heterogeneous High-Low Similarity Selection Function}: with this method we put inside a single Co-Img, a given percentage of pictures with high similarity and a percentage of images with low similarity.
          \end{itemize}
    \item \textbf{Local augmentation}: When balancing all classes in a dataset, the minority class must be amplified to align with the others. To avoid repeating identical patterns, each composite image is slightly rotated, introducing variation while maintaining class balance.
\end{enumerate}
In this paper, we adopt a 3×1 layout format and employ Class-Based Selection Function for grouping input images.

\subsubsection{Description of the Newly Generated Dataset 3x1-Co-OCTDL}
The constructed dataset, referred to as 3x1-Co-OCTDL, consists of newly generated Composite Input Images (Co-Img) designed to enhance both the volume and diversity of the training data. The total number of samples produced is influenced by multiple factors, including the size of the original dataset, the grouping strategy applied, and the selected layout configuration.

\subsection{Class-Based Selection Function}
\subsubsection{Problem definition}
This method constructs Composite Input Images (Co-Imgs) by grouping images belonging to the same class. The generation process follows a structured procedure in which images from a given class are selected and arranged into composite samples of predefined dimensions. The function requires three parameters: the dimensional layout of the Co-Img, the original dataset with its classes, and whether or not intra-compositerepetition of images is permitted.
\begin{align}
    \mathcal{F}(L, \mathcal{D}, r)
     & = \big\{ \mathrm{CoImg}_{ci} \;\big|\;
    \mathrm{CoImg}_{ci} = \mathrm{Compose}(x_{c1}, \ldots, x_{ck}; L), \nonumber  \\
     & \quad i=1,\ldots,N_{c}, \; x_{ck} \in \mathcal{D}_c,\; k = size(L), \; c \in \mathcal{C},\nonumber  \\
     & \quad x_{cj} \neq x_{cp} \text{ only if } r=0, \; j=1,\ldots,k \text{ and } p=1,\ldots,k \big\}
\end{align}
Where
$N_c$ is the size of Co-Img of class $c$ in new dataset ,
$L$ is the layout (e.g., $3 \times 1$),
$\mathcal{D}$ is the original dataset,
$r \in \{0,1\}$ indicates repetition flag, and
$\mathcal{D}_c \subseteq \mathcal{D}$ is the set of samples from class $c$.

The number of images varies across classes, the total number of new dataset is:
\[
    N_{\text{total}} = \sum_{c \in \mathcal{C}} \mathcal{F}(L, \mathcal{D}_c, r)
\]

\subsubsection{Algorithm schema}
\begin{figure}[h!]
    \centering
    \includegraphics[width=0.9\textwidth]{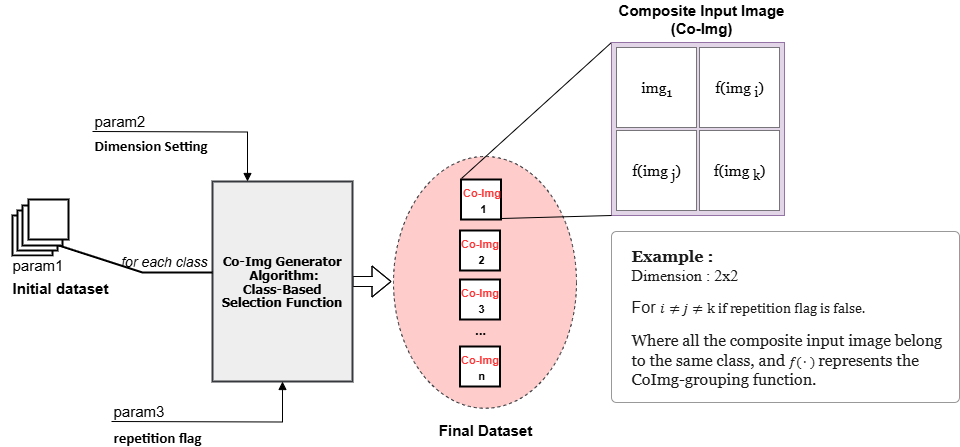}
    \caption{Co-Dataset Generator Algorithm Schema.}
    \label{fig:Constructing_enhanced_schema02}
\end{figure}
\subsubsection{Algorithm Description}

\noindent\textbf{Input:}
\begin{itemize}[labelsep=0.3em, topsep=3pt, partopsep=6pt, parsep=0pt, itemsep=4pt]
    \item A dataset \( D = \{D_c\}_{c \in \mathcal{C}} \) with \( \mathcal{C} \) classes.
    \item Each class \( c \) contains \( N_c = |D_c| \) images.
    \item The desired dimensions of the Co-Img: \( m \times n \). The number of images per Co-Img is \( k = m \times n \).
    \item Repetition flag \( r \in \{0,1\} \):
          \begin{itemize}
              \item \( r=0 \): generate combinations without intra-composite repetition.
              \item \( r=1 \): generate combinations with intra-composite repetition.
          \end{itemize}
\end{itemize}
\noindent---\newline
\noindent\textbf{Output:}
\begin{itemize}[labelsep=0.3em, topsep=3pt, partopsep=6pt, parsep=0pt, itemsep=4pt]
    \item A new dataset \( D_{\text{final}} \), where each Co-Img is a composite image formed by combining \( k \) images from the same class, depending on the repetition flag \( r \).
\end{itemize}
\noindent---\newline
\noindent\textbf{Algorithm Steps:}
\begin{algorithm}[H]
    \caption{Class-Based Selection Function}
    \label{alg:algo1}
    \begin{algorithmic}[1]
        \REQUIRE Dataset \( D \), dimensions \( m \times n \), repetition flag \( r \in \{0,1\} \)
        \ENSURE Enriched dataset \( D_{\text{final}} \)

        \STATE Initialize \( D_{\text{final}} \gets \varnothing \)  \COMMENT{Empty list for CoImg}
        \STATE \( k \gets m \times n \) \COMMENT{Number of images per CoImg}

        \FOR{each class \( c \in D \)}
        \STATE \( N_c \gets |D_c| \)  \COMMENT{Number of images in class \( c \)}

        \IF{\( r = 1 \)}
        \STATE \( \text{Combinations}_c \gets \text{generate\_multicombinations}(D_c, k) \) \COMMENT{with repetition}
        \ELSE
        \STATE \( \text{Combinations}_c \gets \text{generate\_combinations}(D_c, k) \) \COMMENT{without repetition}
        \ENDIF

        \FOR{each combination \( W \in \text{Combinations}_c \)}
        \STATE \( \text{CoImg} \gets \text{create\_CoImg}(W, m, n) \)
        \STATE \( D_{\text{final}} \gets D_{\text{final}} \cup \{\text{CoImg}\} \)
        \ENDFOR
        \ENDFOR

        \RETURN $D_{\text{final}}$
\end{algorithmic}
\end{algorithm}

\noindent\textbf{Helper Functions:}
\paragraph{1. Generate Combinations (without repetition)}
\leavevmode\newline
\texttt{function generate\_combinations(\(S, k\))} \\ 
\indent\textbf{return} all size-\(k\) subsets of \(S\) (no repetition) 

\paragraph{2. Generate Multi-Combinations (with repetition)}
\leavevmode\newline
\texttt{function generate\_multicombinations(\(S, k\))}\\
\indent\textbf{return} all size-\(k\) multisets from \(S\) (with repetition)

\paragraph{3. Create CoImg}
\leavevmode\newline
\texttt{function create\_CoImg(\(W, m, n\))}\\
\indent CoImg \( \gets \) initialize\_grid(\(m, n\)) \\
\indent \textbf{for} \( i = 1 \) \textbf{to} \( m \) \textbf{do} \\
\indent \quad \textbf{for} \( j = 1 \) \textbf{to} \( n \) \textbf{do} \\
\indent \quad \quad CoImg[\(i,j\)] \( \gets \) \( W[(i-1)\cdot n + j] \) \\
\indent \quad \textbf{end for} \\
\indent \textbf{end for} \\
\indent \textbf{return} CoImg

--- 

---
\subsection{Advantages of the new dataset}
By merging multiple images from the same class into a single composite input, weaker images are reinforced by stronger ones. As a result, the newly formed samples contain richer and more relevant information, which significantly supports the training process. This approach effectively mitigates challenges related to small dataset size and low-quality samples, providing the model with more robust and informative inputs.
The main benefits of the Class-
Based Input Image Composition (CB-ImgComp) approach can be summarized as follows:
\subsubsection{Enhancing Variability:}
Each Co-Img combines several input images, producing enriched patterns and variations that consolidate the most relevant visual cues. This fusion augments the informative content of individual samples, enabling the model to extract stronger and more generalizable features for improved learning.
\begin{figure}[h!]
    \centering
    \includegraphics[width=0.75\textwidth]{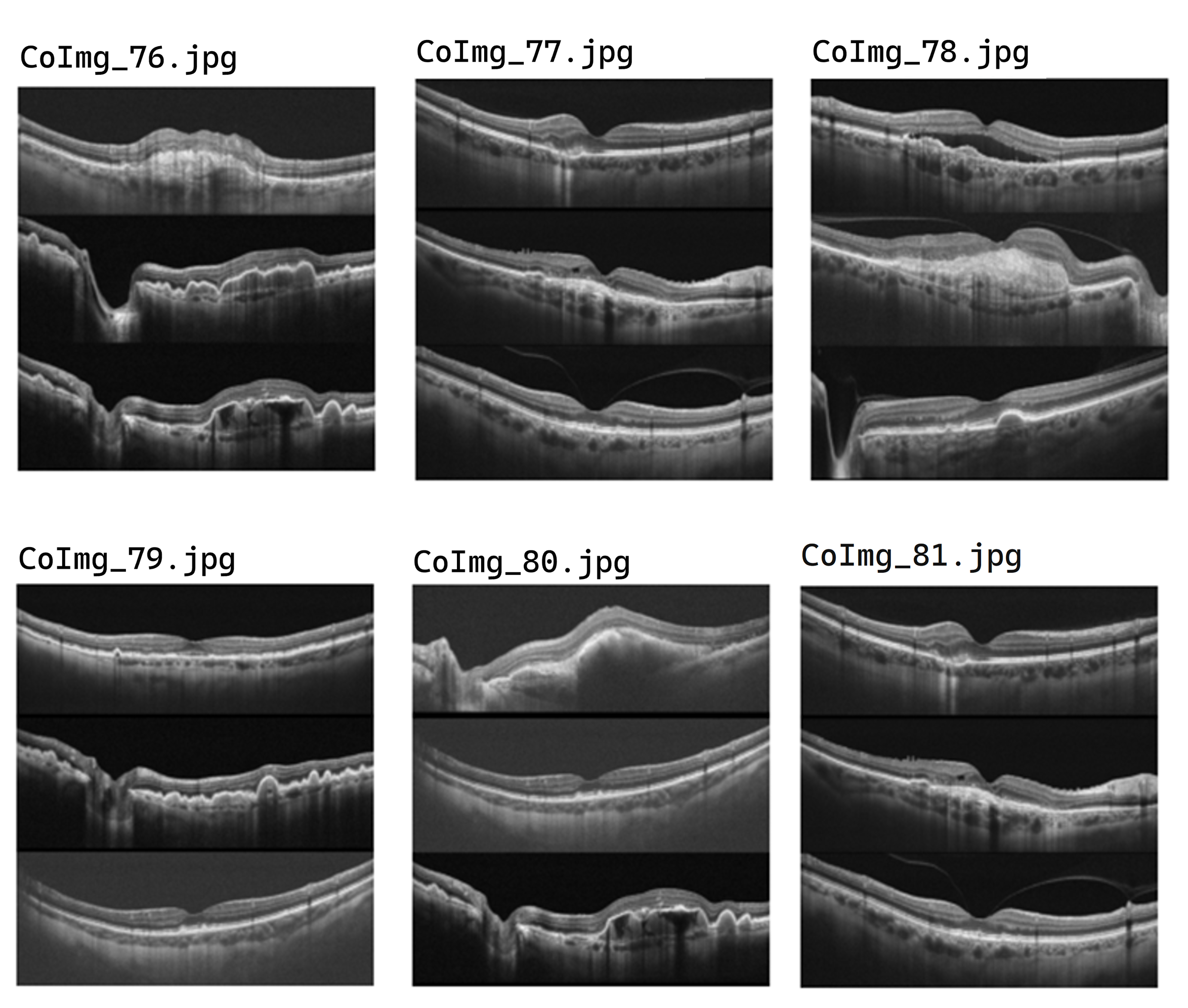}
    \caption{Co-OCDTL AMD Class examples}
    \label{fig:amd_examples}
\end{figure}
\subsubsection{Increasing Dataset Size:}
By merging multiple input images into a single composite image with repetition, the dataset size grows exponentially, depending on the number of possible combinations, thereby providing a larger pool of training examples. To prevent the occurrence of repeated patterns across different composite images, a slight rotation is applied to each constituent image. This ensures that every CoImg remains unique, while enriching variability and reducing the risk of the model memorizing redundant structures.
\begin{figure}[h!]
    \centering
    \includegraphics[width=0.58\textwidth]{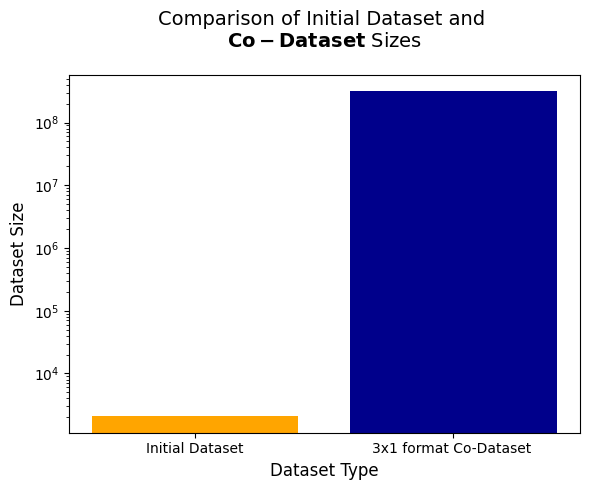}
    \caption{Dataset size expansion using CB-ImgComp generation with $k$=3}
    \label{fig:size_expansion}
\end{figure}
\subsubsection{Reducing False Metrics}
By combining several input images into a single composite, the likelihood of encountering two highly similar images from different classes is reduced. Such similarities are often a source of misclassification and erroneous predictions. Therefore, this strategy helps to minimize false metrics and improves the reliability of the model’s performance.

The proposed dataset construction approach also provides the ability to achieve class balance.
By generating composite images systematically across all categories, the method ensures that no class remains underrepresented.
This balancing capacity addresses one of the critical limitations of many medical imaging datasets, where certain diseases are less frequent than others.
A detailed explanation of how balanced classes are obtained is presented in the next section.

\subsection{Perfect Balancing Class Representation}
After applying composite-image generation, the dataset expands combinatorially. Crucially, the minority class gains a substantial number of informative samples, strengthening its representation. This is because we designate it as the baseline to which all other classes must be aligned.
\subsubsection{Balancing with the Minority Class}
For majority classes, we enforce a no-reuse rule (“if an image is used, don’t use it again”) when forming composites; this naturally caps their composite count while preserving the salient information within each composite.

To achieve balanced classes, we define a target number of composites $T$, corresponding to the expansion obtained for the minority class.
For every other class, composites are generated under the following principle:
\begin{enumerate}[label=--,leftmargin=2em,labelwidth=1em,labelsep=0.3em,align=parleft,topsep=6pt, partopsep=6pt, parsep=0pt, itemsep=4pt]
    \item Each composite contains distinct images (no intra-composite repetition).
    \item If a majority class cannot reach the target $T$ under these constraints,
          the remaining composites are completed by reusing the available samples with
          light augmentations (e.g., small rotations, translations, or contrast adjustments).
\end{enumerate}
This procedure ensures that all classes ultimately reach the same target $T$, thereby producing a perfectly balanced dataset while maintaining
informative and diverse content, as illustrated in Table~\ref{tab:composites_t},
where each disease class (e.g., AMD, DME, etc.) is expanded or reduced up to $T(1540)$ through non-repetitive composites or slight augmentations.

\begin{table}[H]
    \centering
    \caption{Composite Generation Statistics per Retinal Disease Class}
    \label{tab:composites_t}
    \setlength{\tabcolsep}{3pt}
    \renewcommand{\arraystretch}{1.3}
    \begin{tabularx}{\columnwidth}{|l|r|r|X|}
        \hline
        \textbf{Disease} & \textbf{Scans} & \textbf{Co-Scans} & \textbf{New Representation}                                              \\
        \hline
        \textbf{AMD}     & 1 231           & 310 144 295       & \multirow{6}{*}{\begin{minipage}[c]{4cm}%
                                                                                        Majority classes. \newline Expand/Reduce up to T(1 540)
                                                                                    \end{minipage}} \\
        \cline{1-3}
        \textbf{DME}     & 147            & 518 665           &                                                                          \\
        \cline{1-3}
        \textbf{ERM}     & 155            & 608 685           &                                                                          \\
        \cline{1-3}
        \textbf{NO}      & 332            & 6 044 060         &                                                                          \\
        \cline{1-3}
        \textbf{RVO}     & 101            & 166 650           &                                                                          \\
        \cline{1-3}
        \textbf{VID}     & 76             & 70 300            &                                                                          \\
        \hline
        \textbf{RAO}     & 22             & 1 540             & Minority class. Minimal number of samples ($T$=1 540).                    \\
        \hline
        \textbf{Total}   & 2 064           & 317 554 195       & 10 780 ($T\times7$)                                                      \\
        \hline
    \end{tabularx}
\end{table}

\subsubsection{Minority-Class-Driven Balancing Algorithm}
The following algorithm outlines how the minority class is leveraged as a baseline to systematically align all other classes, thereby achieving a perfectly balanced dataset enriched with highly informative, diverse, and discriminative samples.
\begin{algorithm}[H]
    \caption{Balanced Co-Dataset Generation}
    \label{alg:algo2}
    \begin{algorithmic}[1]
        \REQUIRE A dataset  \( D \) with \( N \) classes, where class \( i \) has \( N_i \) images; composite size \( k \) (images per CoImg)
        \ENSURE A balanced dataset where each class has \( T \) CoImg images
        \STATE Use the procedure defined in Algorithm~\ref{alg:algo1} as a CBSFunction
        \STATE \( T \gets \min\limits_{i=1,\ldots,N} \binom{N_i}{k} \) \COMMENT{Baseline: minimum possible \( k \)-combinations across classes}

        \FOR{each class \( C \) with \( N_C \gets |C| \)}
        \STATE \( M \gets \binom{N_C}{k} \) \COMMENT{Total unique \( k \)-combinations in class \( C \)}
        \IF{\( M > T \)}
        \STATE $\mathrm{CBSFunction}({k},D, 0)$
        \COMMENT{Generate without intra-composite repetition.}
        \ELSE
        \STATE $\mathrm{CBSFunction}({k},D, 1)$
        \COMMENT{Generate all unique composites; if $T$ is not reached, complete with slight augmentation.}

        \ENDIF
        \STATE Store the resulting \( T \) composite candidates (Co-Img) for class \( C \)
        \ENDFOR
        \STATE Merge all class-specific \( T \) composite sets into the final balanced dataset
        \RETURN Final balanced dataset
    \end{algorithmic}
\end{algorithm}

This strategy ensures the diversity of each class is maintained without discarding samples to balance the dataset.
\begin{figure}[H]
    \centering
    \includegraphics[width=0.85\textwidth]{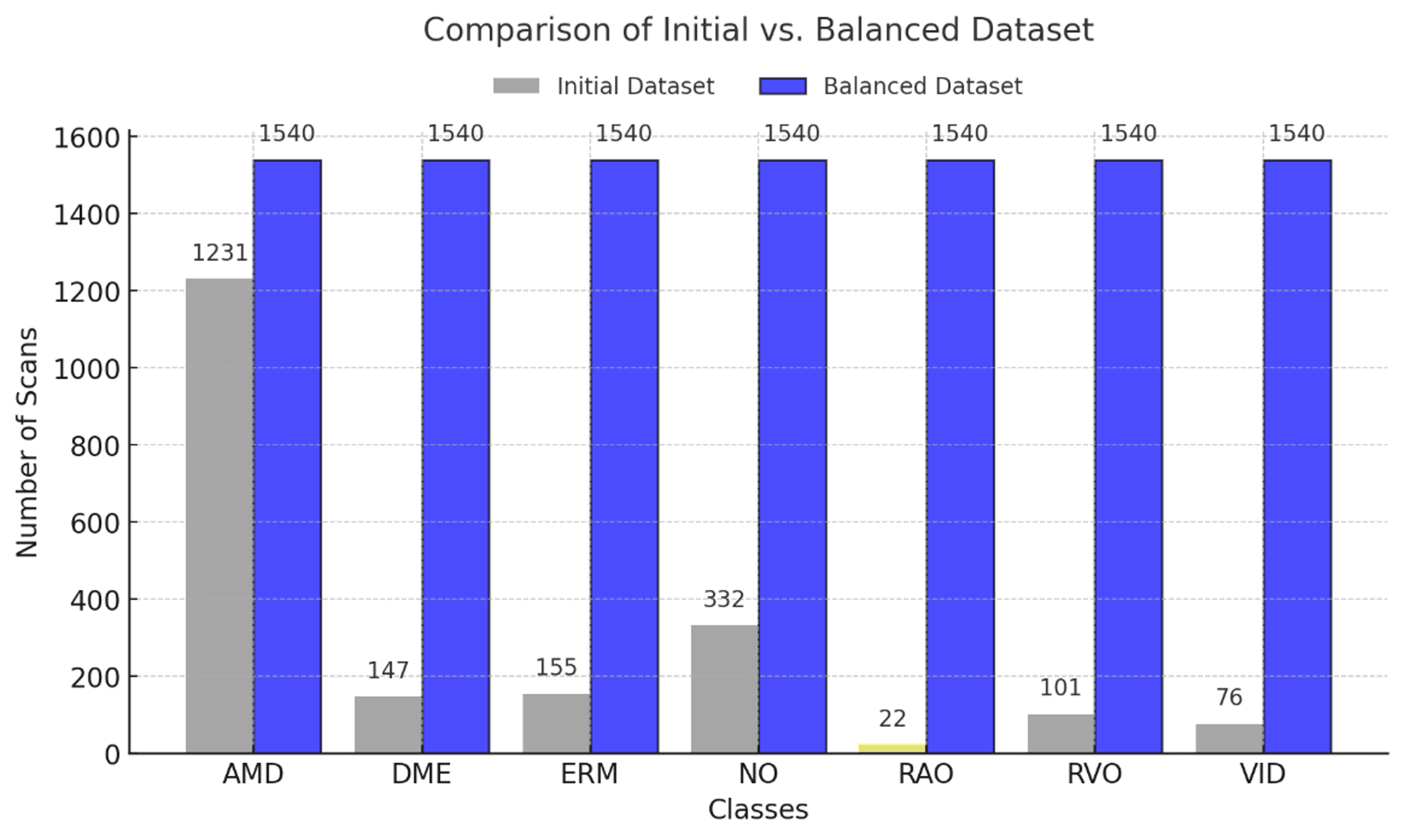}
    \caption{Comparison of Initial Dataset and Balanced Co-Dataset}
    \label{fig:balanced_ds}
\end{figure}

\section{Experiment}
This section details the experimental design used to validate the efficacy of our Class-Based Input Image Composition (CB-ImgComp) approach. The primary objective is to quantitatively measure the performance improvement of a deep learning model when trained on our newly generated, balanced dataset compared to the original, imbalanced dataset.

\subsection{Experimental Setup}
To ensure reproducibility and leverage high-performance computing resources, all experiments were conducted on the Kaggle platform, utilizing its GPU-accelerated notebook environments.

The core of our analysis involves the VGG16 architecture, pre-trained on ImageNet. For a fair comparison, the model architecture and all training hyperparameters were held constant across all experimental scenarios. The only variable was the training data representation. The model's classification head was adapted for our 7-class problem, and the training was configured with the parameters outlined in Table~\ref{tab:hyperparameters} \citep{kulyabin2024octdl}.

\begin{table}[H]
    \centering
    \caption{Consistent Hyperparameters for All Experiments}
    \label{tab:hyperparameters}
    \begin{tabular}{ll}
        \toprule
        \textbf{Parameter} & \textbf{Value}               \\
        \midrule
        Model Architecture & VGG16 (ImageNet pre-trained) \\
        Optimizer          & Adam                         \\
        Learning Rate      & $1 \times 10^{-4}$           \\
        Loss Function      & Categorical Cross-Entropy    \\
        Batch Size         & 32                           \\
        Epochs per Fold    & 50                           \\
        \bottomrule
    \end{tabular}
\end{table}

\subsection{Datasets}
Two dataset configurations were used to conduct a direct comparison:
\begin{enumerate}
    \item \textbf{Baseline Dataset (OCTDL):} The original, imbalanced dataset containing 2,064 individual OCT scans.
    \item \textbf{Proposed Dataset (3x1-Co-OCTDL):} The perfectly balanced dataset generated using our CB-ImgComp method. This dataset consists of 10,780 composite images, each with a 3x1 layout, and is publicly available on Kaggle\footnote{https://www.kaggle.com/datasets/azdineh/c-dataset-2025}.
\end{enumerate}

\subsection{Training and Validation Protocol}
To ensure a robust and unbiased evaluation, a \textbf{5-fold cross-validation} strategy was employed for both the baseline and the proposed datasets. In each fold, the data was partitioned into an 80\% training set and a 20\% validation set.

The experimental procedure was as follows:
\begin{itemize}
    \item \textbf{Baseline Model:} The VGG16 model was trained and evaluated five times using the 5-fold splits of the original OCTDL dataset.
    \item \textbf{Proposed Model:} The identical VGG16 model was trained and evaluated five times using the 5-fold splits of our 3x1-Co-OCTDL dataset.
\end{itemize}
The final performance metrics were calculated by averaging the results from the five folds. This approach minimizes the impact of data partitioning and provides a more reliable measure of the model's generalization capability.

\subsection{Evaluation Metrics}
The performance of each model was assessed using a standard set of classification metrics to provide a comprehensive view of its diagnostic accuracy and reliability. The key metrics include:
\begin{itemize}
    \item Accuracy
    \item Precision (Macro-Averaged)
    \item Recall (Macro-Averaged)
    \item F1-Score (Macro-Averaged)
\end{itemize}
Additionally, to directly measure the model's ability to reduce diagnostic errors, we specifically analyzed the False Positive Rate (FPR) and False Negative Rate (FNR).

\section{Results}
This section presents the comparative performance of the VGG16 model trained on the original, imbalanced OCTDL dataset (Baseline) versus the model trained on our balanced, composite-image dataset, 3x1-Co-OCTDL (Proposed Method). The results demonstrate a substantial improvement in diagnostic accuracy and reliability across all evaluation metrics.

\subsection{Quantitative Performance Metrics}
The overall performance of the two models is summarized in Table~\ref{tab:performance_metrics}. The statistics show unequivocally that the proposed approach, utilizing our CB-ImgComp method, far exceeds the baseline across all assessment criteria.

The model trained on the Co-OCTDL dataset achieved near-perfect scores, with an \textbf{accuracy of 99.70\%}, a \textbf{macro F1-Score of 0.997}, and \textbf{Precision and Recall values of 0.997}. This represents a significant improvement over the baseline model, which scored 85.9\% on accuracy and 0.869 on F1-Score. Furthermore, the \textbf{AUC score} saw a remarkable increase from \textbf{0.977 to 0.9996}, indicating that the new model has an exceptionally high capability to distinguish between the different retinal disease classes. This dramatic performance gain underscores the robustness and reliability of the model when trained on the perfectly balanced and feature-enriched composite dataset.

\begin{table}[H]
    \centering
    \caption{Comparison of Performance Metrics between the Baseline and Proposed Method.}
    \label{tab:performance_metrics}
    \begin{tabular}{lccccc}
        \toprule
        \textbf{Method}               & \textbf{Accuracy} & \textbf{F1-Score} & \textbf{AUC} & \textbf{Precision} & \textbf{Recall} \\
        \midrule
        Baseline (VGG16 - OCTDL)      & 0.859             & 0.869             & 0.977        & 0.888              & 0.859           \\
        CB-ImgComp (VGG16 - Co-OCTDL) & 0.9970            & 0.9970            & 0.9996       & 0.9971             & 0.9970          \\
        \bottomrule
    \end{tabular}
\end{table}

\subsection{Analysis of Misclassification Errors}
\begin{figure}[H]
    \centering
    \includegraphics[width=\columnwidth]{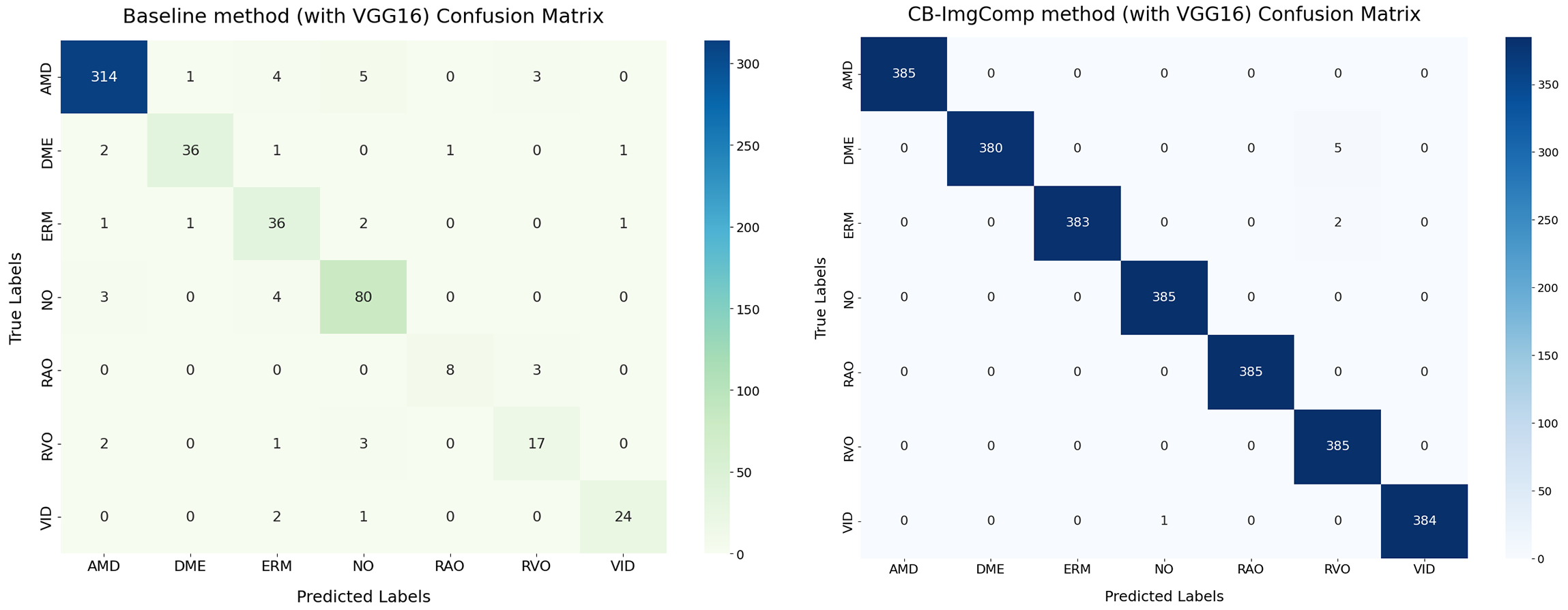}
    \caption{Confusion matrices for the classification of retinal diseases. Left:The baseline model (Classical Method) shows significant off-diagonal misclassifications.
        Right: The model trained with our proposed CB-ImgComp method shows a drastic reduction in classification errors, with a nearly perfect diagonal.}
    \label{fig:confusion_matrices}
\end{figure}

To further analyze the model's performance on a per-class basis, we generated confusion matrices for both the baseline and the proposed method, as shown in Figure~\ref{fig:confusion_matrices}.

The confusion matrix for the baseline model (left) reveals several misclassifications, particularly among classes with high visual similarity. For instance, Retinal Vein Occlusion (RVO) and Age-related Macular Degeneration (AMD) were frequently confused with other pathologies. In contrast, the confusion matrix for the model trained with our CB-ImgComp method (right) displays a nearly perfect diagonal structure. This visualization confirms a substantial reduction in misclassification errors, with minimal false positives and false negatives across all seven classes. The improved model demonstrates a clear ability to overcome the visual ambiguities that limited the baseline's performance, validating our approach's effectiveness in enhancing discriminative feature learning.

\section{Discussion}
The experimental results unequivocally demonstrate the superiority of our Class-Based Input Image Composition (CB-ImgComp) approach over the baseline method. The notable gains in accuracy, F1-score, and precision are not just incremental improvements but signify a fundamental enhancement in the model's ability to generalize and perform consistently. The near-perfect confusion matrix (Figure~\ref{fig:confusion_matrices}), which shows a drastic reduction in misclassifications, verifies the exceptional reliability of the model when trained on the composite-image dataset. In contrast, the baseline model exhibited higher variability and error rates, particularly in underrepresented classes, indicating a sensitivity to data imbalance and a limited capacity for generalization.

\subsection{Clinical Relevance and Impact}
This performance gap highlights the critical importance of optimizing data representation in sensitive applications such as biomedical analysis. In domains handling health-related data, diagnostic errors can have profound consequences. The dramatic increase in precision and recall achieved with our method corresponds to a significant reduction in both false positives and false negatives. This is a vital step toward building robust and trustworthy AI systems that can be safely integrated into clinical workflows, where diagnostic reliability is paramount.

\subsection{Implications of Findings}

\subsubsection{Addressing Small and Imbalanced Datasets}
A key finding of this study is that exceptional accuracy can be achieved even when the initial dataset is small and severely imbalanced. Our CB-ImgComp method serves as a powerful form of semantic data augmentation and balancing. By creating perfectly balanced, feature-rich composite samples, we directly address the challenges that plague many medical imaging datasets. While the baseline model struggled with minority classes, our approach ensures consistent and high-fidelity classification across all categories by enriching the training data at the input level, rather than relying solely on algorithmic modifications like weighted loss functions.

\subsubsection{Minimizing False Diagnostic Results}
In clinical practice, minimizing false positives (which can lead to unnecessary treatments) and false negatives (which can lead to missed diagnoses) is absolutely critical. The proposed method's ability to reduce these errors to near-zero levels suggests its potential for safe and effective real-world application. This success stems from providing the model with a more comprehensive and less ambiguous representation of each pathology, thereby enhancing its discriminative power and reducing diagnostic uncertainty.



\section{Conclusion}
In this paper, we addressed the critical challenge of training accurate deep learning models on small and imbalanced medical datasets. We introduced \textbf{Class-Based Input Image Composition (CB-ImgComp)}, a novel input-level preprocessing technique designed to create a perfectly balanced and feature-enriched dataset by composing multiple same-class images into single, information-dense training samples.

Our experimental evaluation on the OCTDL dataset demonstrated the profound effectiveness of this approach. By training a standard VGG16 model on our generated \textbf{Co-OCTDL} dataset, we achieved near-perfect diagnostic performance, with an accuracy of 99.7\% and an AUC of 0.9996. This represents a substantial improvement over the baseline model and, crucially, resulted in a drastic reduction of the false positive and false negative rates that often hinder the clinical adoption of AI models.

The primary contribution of this work is a simple, model-agnostic, and highly effective strategy that enhances data quality at its source. Our findings confirm that intelligent data representation can be more impactful than relying on complex model architectures alone. We conclude that the CB-ImgComp method holds significant promise for developing more robust, reliable, and trustworthy AI-driven diagnostic tools, particularly in medical fields where data scarcity and imbalance are persistent obstacles.

\bibliographystyle{apalike}
\bibliography{mybiblio}

\end{document}